\documentclass[runningheads]{llncs}

\usepackage{eccv}

\usepackage{eccvabbrv}

\usepackage{graphicx}
\usepackage{booktabs}

\usepackage[accsupp]{axessibility}  

\usepackage{amsmath}
\usepackage{amssymb}
\usepackage{mathtools}
\usepackage{amsfonts}
\usepackage{tabularx}
\usepackage{booktabs} 
\usepackage[table]{xcolor} 
\usepackage{multirow}

\usepackage{hyperref}

\usepackage{orcidlink}

\begin{document}

\title{RotateAttention: RoPE-Aware Rotation and Range Rectification for INT4 Quantized Attention in Video Generation} 

\titlerunning{RotateAttention}





\newcommand{\equalcontrib}{\textsuperscript{\(\ast\)}}
\newcommand{\corr}{\textsuperscript{\(\dagger\)}}

\author{
Yaofu Liu\inst{1}\equalcontrib \and
Wanli Lan\inst{2}\equalcontrib \and
Jinxi Li\inst{2}\equalcontrib \and
Binhang Yuan\inst{1}\corr \and
Harry Yang\inst{1}\corr
}

\authorrunning{Y.~Liu et al.}

\institute{
The Hong Kong University of Science and Technology, Hong Kong SAR, China\\
\email{yliuls@connect.ust.hk, biyuan@ust.hk, yangharry@ust.hk}
\and
Independent Researcher, China
}

\maketitle

\begingroup
\renewcommand\thefootnote{\(\ast\)}
\footnotetext{Equal contribution.}
\renewcommand\thefootnote{\(\dagger\)}
\footnotetext{Corresponding authors.}
\endgroup

\begin{abstract}
In \textbf{DiT-based video generation models equipped with 3D Rotary Position Embeddings (3D RoPE)}, the attention mechanism remains a primary computational bottleneck due to its quadratic complexity with respect to sequence length. While quantized \textbf{FlashAttention} offers a promising path toward hardware acceleration, existing low-bit quantization methods overlook two critical challenges in this setting: \textbf{1)} applying online rotation matrices---a widely used technique for mitigating outliers in Queries ($Q$) and Keys ($K$)---is difficult to reconcile with \textbf{RoPE}; and \textbf{2)} the non-negative attention matrix $P = \exp(QK - \max(QK))$ makes symmetric quantization waste half of the 4-bit dynamic range. In this work, we observe that the outlier distributions of $Q$ and $K$ are strongly affected by the dimensional partitioning of \textbf{3D RoPE}. Based on this finding, we propose \textbf{RotateAttention}, an efficient \textbf{mixed-precision INT4 FlashAttention} framework tailored for \textbf{DiT-based video generation models with 3D RoPE}, using selective \textbf{FP16 fallback} for accuracy-sensitive attention blocks and denoising steps. RotateAttention introduces two core techniques: \textbf{1) RoPE-aware Rotation}, which employs either mergeable rotation matrices that can be fused into RoPE or negligible-overhead matrices to mitigate RoPE-induced outliers in $Q$ and $K$; and \textbf{2) Range-optimized $P$ Quantization}, which uses fixed scales and zero-points to fully exploit the \textbf{INT4 numerical range} with minimal computational overhead. 
Experiments show that \textbf{RotateAttention} preserves video generation quality nearly identical to full-precision baselines while achieving up to 1.68$\times$ end-to-end speedup and 2.2$\times$ kernel-level acceleration.

\end{abstract}

\section{Introduction}

Diffusion models~\cite{peebles2023dit} have achieved remarkable success in video generation; however, they face a severe computational bottleneck within the attention mechanism. Since video generative models often involve sequences exceeding $10\text{K}$ tokens, the attention cost scales quadratically with the token count. A promising approach to mitigate this is 4-bit quantization of FlashAttention, a method that is orthogonal to other strategies such as sparse attention\cite{zhang2025spargeattention,yang2025sparse,zhang2025sla,xi2025sparse,liu2021swin,chu2021twins,xiao2023efficient,xiao2024infllm,chen2024longlora,jiang2024minference,venkataramanan2024skipattention,yuan2024ditfastattn, zhang2025fast}. We develop a \textbf{highly compatible mixed-precision INT4 attention framework} for DiT-based video generation models with \textbf{3D-RoPE},
substantially improving quantization accuracy by selectively falling back to \textbf{FP16} in accuracy-sensitive attention blocks and denoising steps.

\paragraph{\textbf{Outliers and Rotation Transformation}}
The accuracy of low-bit quantized attention is constrained by the presence of \textit{outliers} in the tensors---elements whose magnitudes deviate significantly from the mean distribution. These extreme values compress the dynamic range available for the majority of the tensor, leading to severe precision loss. Rotation transformations are established techniques to mitigate outliers by redistributing tensor values, as discussed in detail in \cref{sec:relate}.

\paragraph{\textbf{Limitations of Existing Work.}}
Current 4-bit attention methods, when applied to DiT-based video generation models with 3D RoPE, face several limitations:
1) The Query ($\mathbf{Q}$) and Key ($\mathbf{K}$) matrices are usually not rotated, although rotation can effectively equalize outliers. This is because common rotation matrices are difficult to fuse with Rotary Position Embeddings (RoPE)~\cite{su2024roformer}, while online rotation introduces non-negligible overhead.
2) Existing methods employ symmetric quantization for the probability matrix $\mathbf{P} = \exp(\mathbf{S} - m)$. Since $\mathbf{P} \in (0, 1]$, this leaves half of the INT4 representation range unused.
Furthermore, LLM-oriented quantized attention methods with $\mathbf{Q}$/$\mathbf{K}$ rotations do not directly transfer to mixed-precision INT4 attention in 3D-RoPE video DiTs.

\paragraph{\textbf{Motivation and Key Findings.} }
In this work, we make the following observations: 
1) We investigate the impact of 3D RoPE on the outlier distribution of $\mathbf{Q}$ and $\mathbf{K}$ in 3D-RoPE video DiTs. We identify specific patterns in the outlier distribution that can be exploited to implement a highly efficient and effective rotation matrix. 
2) Fully utilizing the 4-bit numerical range for the quantization of $\mathbf{P}$ yields a significant improvement in attention accuracy. 

\paragraph{\textbf{Method.} }
We introduce \textbf{RotateAttention}, which comprises two core innovations complemented by a standard offline technique: 
1) \textbf{RoPE-aware Rotation:} A block-diagonal matrix composed of $2 \times 2$ orthogonal blocks that effectively mitigates outliers in $\mathbf{Q}$ and $\mathbf{K}$ with minimal overhead, designed to be compatible with or mergeable into RoPE. 
2) \textbf{Range-Optimized $\mathbf{P}$ Quantization:} A fixed-scale and zero-point calibration for matrix $\mathbf{P}$ that fully exploits the INT4 dynamic range at low computational cost. 
Additionally, we apply a \textbf{Hadamard Transform for $\mathbf{V}$}, a well-established technique~\cite{liu2025spinquant,sun2025flatquant} that mitigates outliers in $\mathbf{V}$ by fusing Hadamard matrices into the existing linear projections at zero inference cost.

Experimental results demonstrate that our method is both efficient and effective. More importantly, our strategy can be integrated into other quantized attention frameworks to further enhance their performance.

\section{Related Work}
\label{sec:relate}

Recent advances in quantization \cite{dettmers2022gpt3int8, frantar2022gptq, lin2023awq, li2024svdquant, van2025fptquant, tao2025plug, li2025mbq, li2024evaluating, zhao2025viditq, shang2023post, li2023q, zhao2024mixdq, lin2024duquant, huang2025qvgen,tang2024post,liu2024eda,sui2024bitsfusion} have significantly improved model efficiency. For matrix multiplication $\mathbf{Y} = \mathbf{X}\mathbf{W}^\top$, where $\mathbf{X} \in \mathbb{R}^{k \times n}$ and $\mathbf{W} \in \mathbb{R}^{m \times n}$, common strategies for maintaining high accuracy include:

\paragraph{1) Per-channel Scaling.} This method\cite{xiao2022smoothquant} mitigates outliers by applying a diagonal scaling matrix $\text{diag}(\mathbf{c})$ to transfer outlier energy from $\mathbf{X}$ to $\mathbf{W}$: $\mathbf{Y} = (\mathbf{X} \text{diag}(\mathbf{c})^{-1}) \cdot (\text{diag}(\mathbf{c})\mathbf{W}^\top)$.

\paragraph{2) Rotation Transformation.} A more general approach uses a transformation matrix $\mathbf{H}$ and its inverse transpose $\mathbf{H}^{-\top}$ such that $\mathbf{Y} = (\mathbf{X}\mathbf{H})(\mathbf{W}\mathbf{H}^{-\top})^\top$. Orthogonal rotations (e.g., Hadamard)\cite{ashkboos2024quarot} redistribute outlier energy across channels to reduce maximum values within each matrix (\textit{intra-matrix}). Non-orthogonal invertible transforms act as a composite of rotation and scaling.

\paragraph{\textbf{Low-bit Attention in Video Generation.}} Current low-bit video attention methods focus on quantizing $\mathbf{Q}\mathbf{K}^\top$ and $\mathbf{P}\mathbf{V}$ using alternative techniques:
(1) \textbf{QKV Smoothing:} The SageAttention series \cite{zhang2025sageattention1, zhang2025sageattention2} uses INT4/INT8 formats, employing channel smoothing (e.g., $\mathbf{Q} - \mathbb{E}[\mathbf{Q}]$) to suppress biased outliers. SageAttention3~\cite{zhang2025sageattention3} further extends this line with NVFP4 micro-scaling formats.
(2) \textbf{Token Reordering:} PAROAttention \cite{zhao2025paroattention} reorders tokens in $\mathbf{Q}$, $\mathbf{K}$, and $\mathbf{V}$ so that each block of $\mathbf{P}$ contains values with similar magnitudes, thereby improving per-block quantization precision. This token-level strategy is \emph{complementary} to our channel-level rotation approach and can potentially be combined for further gains.

\paragraph{\textbf{Low-bit Attention in LLMs.}} LLM quantization typically targets the memory-bound KV cache \cite{liu2026pmkvq}. To mitigate quantization error, rotation-based methods like FlatQuant and SpinQuant\cite{sun2025flatquant,liu2025spinquant} utilize invertible transforms, while \textbf{QuaRot} \cite{ashkboos2024quarot} employs Hadamard matrices. Recent advances, such as FlashAttention-3 \cite{shah2024flashattention}, further optimize efficiency by reducing rotation overhead within high-performance kernels.

\begin{figure*}[t]
    \centering
    \includegraphics[width=1\linewidth]{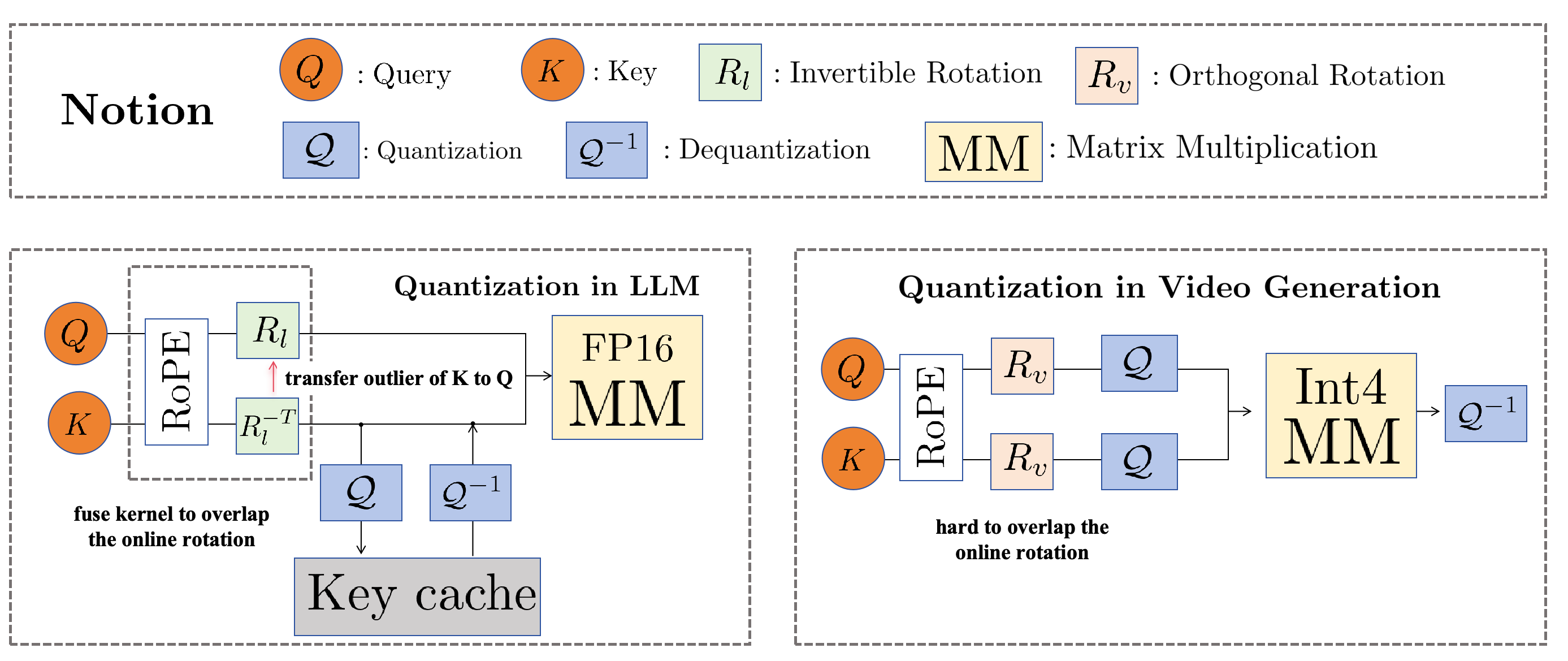}
    \caption{\textbf{Quantization differences between LLM decoding and 3D-RoPE video DiTs}. In LLMs, rotations can prioritize $\mathbf{K}$ quantization and transfer part of the outlier burden to $\mathbf{Q}$. In 3D-RoPE video DiTs, both $\mathbf{Q}$ and $\mathbf{K}$ must be quantized simultaneously, which requires orthogonal transforms and low-overhead execution.}
    \label{fig:llmvsd}
\end{figure*}

\paragraph{\textbf{Key Differences: LLM vs. 3D-RoPE Video DiTs.}}
Fig.~\ref{fig:llmvsd} illustrates the key differences between LLM decoding and 3D-RoPE video DiT attention, highlighting why rotation strategies designed for LLM KV-cache quantization do not directly transfer to DiT-based video generation models with 3D RoPE.

We identify two fundamental distinctions:
(1) \textbf{Computational Intensity:} LLM decoding is memory-bound~\cite{patel2024splitwise,qin2024mooncake,zhong2024distserve}, allowing online rotations to overlap with RoPE or attention kernels. In contrast, 3D-RoPE video DiTs are compute-bound, making online rotation overhead significant and difficult to hide.
(2) \textbf{Quantization Symmetry:} LLM quantization often keeps $\mathbf{Q}$ in FP16 and only quantizes $\mathbf{K}$ for the KV-cache. Conversely, 3D-RoPE video DiTs require simultaneous quantization of $\mathbf{Q}$ and $\mathbf{K}$ to leverage low-bit matmul hardware for actual speedup.

Furthermore, we extend Hadamard transforms to the Value ($\mathbf{V}$) matrix. While previously used in LLM KV-caches, we formally prove their compatibility with smoothing for $\mathbf{V}$. By bridging these rotation strategies and enhancing $\mathbf{P}$ quantization, \textbf{RotateAttention} significantly improves mixed-precision INT4 attention quality for 3D-RoPE video DiTs.

\section{RoPE Effect on Outlier Distribution}\label{sec:outlier_analy}

Let $\mathbf{Q}$, $\mathbf{K}$, and $\mathbf{V}$ denote the query, key, and value matrices, respectively, each belonging to $\mathbb{R}^{N \times D}$, where $N$ represents the sequence length and $D$ denotes the feature dimension.

In DiT-based video generation models, 3D RoPE is widely adopted to capture complex spatiotemporal dependencies. Theoretically, the feature dimension $D$ of the $\mathbf{Q}$ and $\mathbf{K}$ matrices is partitioned into \textbf{three distinct segments}---$D_{f}$, $D_{h}$, and $D_{w}$---corresponding to the temporal (frame), height, and width dimensions, respectively, such that $D=D_{f}+D_{h}+D_{w}$.

The 3D RoPE rotary matrix $\mathbf{M} \in \mathbb{R}^{D \times D}$ is a block-diagonal matrix 
composed of $D/2$ fundamental rotation blocks $\mathbf{M}_i \in \mathrm{SO}(2)$. 
This matrix is partitioned into three sub-matrices—$\mathbf{M}^f$, $\mathbf{M}^h$, and 
$\mathbf{M}^w$—each responsible for encoding temporal, height, and width information 
within their respective dimensional segments.

\paragraph{\textbf{Implementation.}} 
Due to the inherent sparsity of the rotary matrix $\mathbf{M}$, the computation of RoPE substitutes the standard matrix multiplication with a more efficient element-wise operation.

\paragraph{\textbf{Outlier Metric.}} 
To analyze the distribution of outliers that predominantly affect quantization precision, we introduce the \textit{incoherence} metric $\Psi(\mathbf{X})$, which captures the maximum deviation within each channel. For a given matrix $\mathbf{X} \in \{\mathbf{Q}, \mathbf{K}\}$, we first compute the sequence-wise mean vector $\mathbb{E}[\mathbf{X}] \in \mathbb{R}^{1 \times D}$:
\begin{equation}
    \mathbb{E}[\mathbf{X}] = \frac{1}{N} \sum_{i=1}^{N} \mathbf{X}_{i,:},
\end{equation}
where $\mathbf{X}_{i,:}$ denotes the $i$-th row of $\mathbf{X}$. The incoherence $\Psi(\mathbf{X}) \in \mathbb{R}^{1 \times D}$ is then defined as:
\begin{equation}
    \Psi(\mathbf{X}) = \max_{i \in \{1,\dots,N\}} \left| \mathbf{X}_{i,:} - \mathbb{E}[\mathbf{X}] \right|,
\end{equation}
where the absolute value and maximum operations are applied element-wise. This metric provides a channel-level statistical profile, reflecting the outlier concentration within $\mathbf{Q}$ and $\mathbf{K}$ that poses challenges for fixed-point quantization.

\subsection{Data Distribution of $\mathbf{Q}$ and $\mathbf{K}$}

\begin{figure*}[t]
  \centering
  \begin{subfigure}{0.48\linewidth}
    \centering
    \includegraphics[width=\textwidth]{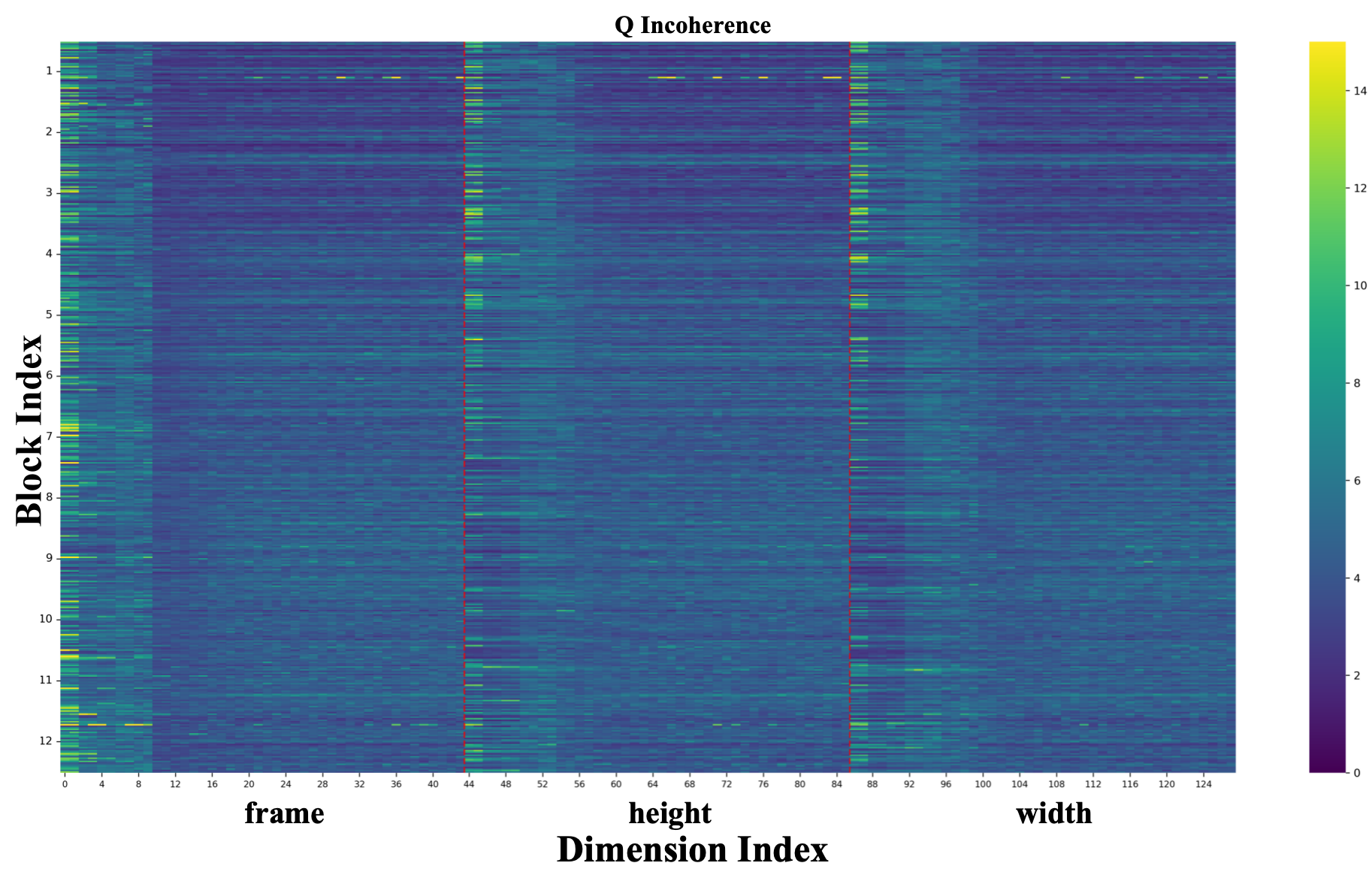}
    \caption{$\mathbf{Q}$ Incoherence}
  \end{subfigure}
  \hfill
  \begin{subfigure}{0.48\linewidth}
    \centering
    \includegraphics[width=\textwidth]{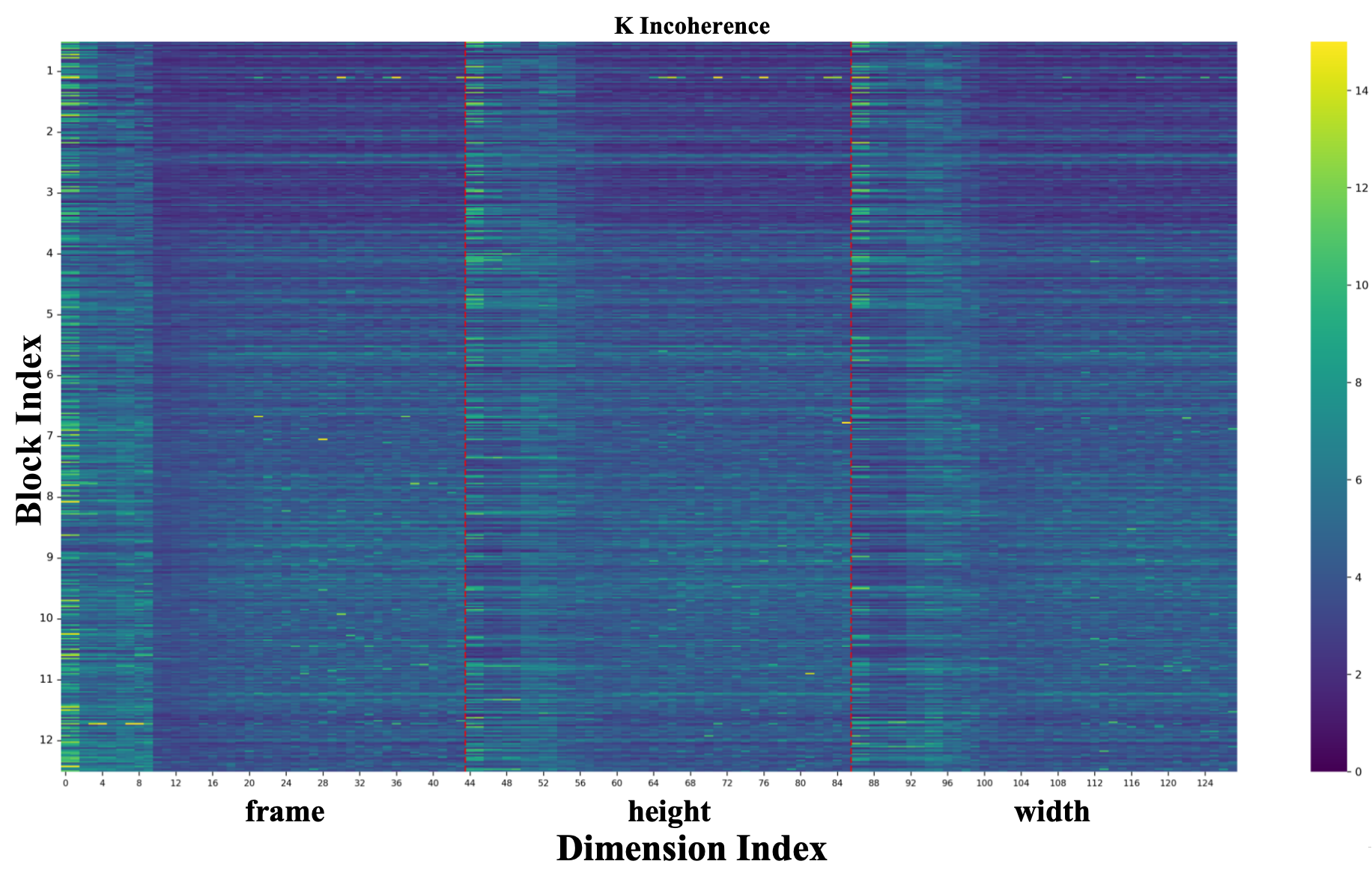}
    \caption{$\mathbf{K}$ Incoherence}
  \end{subfigure}

  \vspace{0.8em}

  \caption{\textbf{Channel-wise incoherence profiles of $\mathbf{Q}$ and $\mathbf{K}$ for Wan2.2.} Incoherence is partitioned into three regions corresponding to the 3D RoPE spatiotemporal segments. We observe strong distributional symmetry between $\mathbf{Q}$ and $\mathbf{K}$, and structured sparsity within each segment.}
  \label{fig:incoherent}
\end{figure*}

By evaluating the incoherence $\Psi(\mathbf{X})$ for $\mathbf{X} \in \{\mathbf{Q}, \mathbf{K}\}$, as shown in Fig.~\ref{fig:incoherent}, we identify two key patterns in the outlier distribution:

\paragraph{\textbf{Spatiotemporal Segmental Incoherence.}} 
Outliers are strictly partitioned into three specific segments---frame, height, and width---matching the channel segments defined by the 3D RoPE. Within these segments, we observe that one half of the dimensions typically contains the vast majority of the incoherence, while the remaining half remains relatively smooth.

\paragraph{\textbf{Cross-Matrix Symmetry.}} 
We find that the outlier distribution patterns of $\mathbf{Q}$ and $\mathbf{K}$ exhibit high symmetry. Specifically, channels at identical indices in $\mathbf{Q}$ and $\mathbf{K}$ generally possess nearly equivalent incoherence magnitudes.

\paragraph{\textbf{Theoretical Interpretation.}}
The segmented outlier structure is consistent with 3D RoPE frequency assignment: low-frequency channel pairs accumulate larger angular spread over long sequences and thus higher variance, while high-frequency pairs remain comparatively smooth. Because $\mathbf{Q}$ and $\mathbf{K}$ share RoPE parameters and are projected from the same hidden states, their channel-wise outlier profiles remain symmetric.

\section{Method}
We focus on attention modules in DiT-based video generation models equipped with 3D RoPE. 
We consider queries ($\mathbf{Q}$), keys ($\mathbf{K}$), and values ($\mathbf{V}$), which are equipped with biases shared across all tokens. To address this, we first perform smoothing on $\mathbf{Q}$, $\mathbf{K}$, and $\mathbf{V}$. 

Furthermore, we observe that $\mathbf{Q}$ and $\mathbf{K}$ exhibit specific distributional patterns of outliers. Based on these patterns, we apply a specially designed rotation for the Query and Key matrices to redistribute and equalize these outliers, thereby improving quantization robustness.

\subsection{Rotation for Query and Key}

\paragraph{\textbf{Outlier Equalization via Orthogonal Rotation.}} 
Motivated by these observations, we introduce an orthogonal rotation matrix $\mathbf{R} \in \mathbb{R}^{D \times D}$ to equalize the outlier distribution. We define $\mathbf{R}$ as a block-diagonal matrix:
\begin{equation}
    \mathbf{R} = \mathrm{diag}\left( \{\mathbf{R}_i\}_{i=1}^{D/2} \right),
\end{equation}
where each $\mathbf{R}_i \in \mathrm{O}(2)$ is a $2 \times 2$ orthogonal matrix. When the rotation matrix $\mathbf{R}$ is non-trainable, we initialize the $2 \times 2$ orthogonal blocks $\mathbf{R}_i$ with a $2 \times 2$ Hadamard matrix, defined as:
\begin{equation}
\label{eq:H_2}
\mathbf{H}_{2} = \frac{1}{\sqrt{2}}
\begin{bmatrix}
 1 &  1 \\
 1 & -1
\end{bmatrix}.
\end{equation}

\subsubsection{Mathematical Analysis on the Necessity of Orthogonality}
\label{sssec:orthogonal_property}

To maintain computational equivalence, when a transformation matrix $\mathbf{R}$ is applied to $\mathbf{Q}$, the matrix $\mathbf{K}$ must be transformed by $\mathbf{R}^{-\top}$:
\begin{equation}
    \mathbf{Q}\mathbf{K}^{\top}
    =
    (\mathbf{Q}\mathbf{R})(\mathbf{K}\mathbf{R}^{-\top})^{\top}.
\end{equation}
This equality holds for any invertible $\mathbf{R}$ in full precision. However, under low-bit quantization, the transformation not only changes the coordinate system, but also directly affects the numerical ranges and outlier distributions of the transformed $\mathbf{Q}$ and $\mathbf{K}$. Since $\mathbf{Q}$ and $\mathbf{K}$ usually exhibit symmetric and comparable outlier patterns after RoPE, the transformation should preserve their balance rather than suppressing outliers in one matrix while amplifying them in the other.

Let $\mathbf{R}=\mathbf{U}\mathbf{\Sigma}\mathbf{L}^{\top}$ be the SVD of $\mathbf{R}$, where $\mathbf{U},\mathbf{L}\in\mathbb{O}(D)$ and $\mathbf{\Sigma}=\mathrm{diag}(\sigma_1,\ldots,\sigma_D)$. The corresponding transformation on $\mathbf{K}$ is
\begin{equation}
    \mathbf{R}^{-\top}
    =
    ((\mathbf{U}\mathbf{\Sigma}\mathbf{L}^{\top})^{-1})^{\top}
    =
    \mathbf{U}\mathbf{\Sigma}^{-1}\mathbf{L}^{\top}.
\end{equation}
Therefore, the singular values applied to $\mathbf{K}$ are $\{1/\sigma_i\}_{i=1}^{D}$. If $\sigma_i<1$, the $i$-th principal direction of $\mathbf{Q}$ is compressed, but the corresponding direction of $\mathbf{K}$ is amplified by $1/\sigma_i>1$. Conversely, if $\sigma_i>1$, the outliers in $\mathbf{Q}$ are amplified while those in $\mathbf{K}$ are suppressed. In both cases, anisotropic scaling introduces an asymmetric range change between $\mathbf{Q}$ and $\mathbf{K}$, which can enlarge the quantization error of either matrix.

To avoid this cross-matrix amplification and preserve balanced quantization difficulty, the transformation should impose the same singular scaling on both branches. This requires $\mathbf{\Sigma}=\mathbf{\Sigma}^{-1}$, i.e., $\sigma_i=1$ for all $i$. Consequently, $\mathbf{\Sigma}=\mathbf{I}$ and $\mathbf{R}=\mathbf{U}\mathbf{L}^{\top}$, which is an orthogonal matrix. This analysis explains why orthogonal rotations are particularly suitable for the symmetric outlier patterns induced by RoPE, as shown in Fig.~\ref{fig:incoherent}.

\subsubsection{Interleaved Rotation (Merge-able)}
\label{sssec:interleaved_rotation}

\begin{figure*}[t]
    \centering
    \includegraphics[width=1\linewidth]{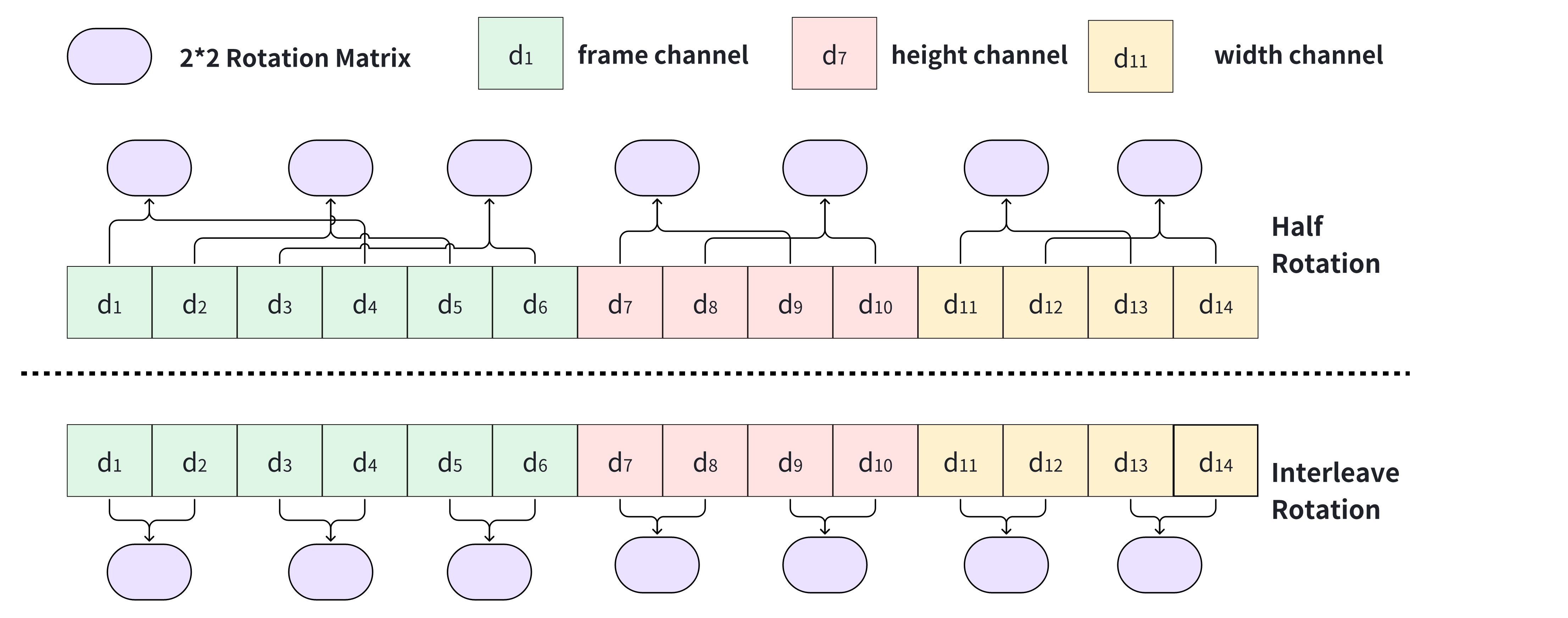}
    \caption{\textbf{Rotation strategies for $\mathbf{Q}$ and $\mathbf{K}$}. Interleaved Rotation groups adjacent dimensions into $2 \times 2$ orthogonal blocks, while Half Rotation pairs each dimension with its counterpart in the other half-segment.}
    \label{fig:rotation_strategies}
\end{figure*}

To integrate with RoPE, we design \textit{Interleaved Rotation}. We define a block-diagonal orthogonal matrix $\mathbf{R}$, where each $2 \times 2$ block acts on adjacent channel pairs $(d_{2i}, d_{2i+1})$ for $i \in \{1, \dots, D/2\}$. This transformation redistributes incoherence, yielding a more uniform outlier profile across the feature dimensions of $\mathbf{Q}$ and $\mathbf{K}$.

The core advantage of this design is its \textbf{operator mergeability}. Since both the rotary matrix $\mathbf{M}$ and our rotation $\mathbf{R}$ consist of $2 \times 2$ orthogonal blocks, their product $\mathbf{M}' = \mathbf{R}\mathbf{M}$ (or $\mathbf{M}\mathbf{R}$) preserves the block-diagonal structure. Consequently, the operations can be fused:
\begin{equation}
    \text{RoPE}(\mathbf{Q})\mathbf{R} = \mathbf{Q}(\mathbf{M}\mathbf{R}), \quad \text{RoPE}(\mathbf{K})\mathbf{R} = \mathbf{K}(\mathbf{M}\mathbf{R}).
\end{equation}
Because $\mathbf{M}\mathbf{R}$ maintains the same sparsity pattern as $\mathbf{M}$, it remains fully compatible with standard element-wise RoPE kernels. By pre-fusing $\mathbf{R}$ into the RoPE weights, we achieve outlier redistribution with \textbf{zero inference overhead}, providing a "free" improvement to quantization precision.

\subsubsection{Half Rotation}
\label{sssec:half_rotation}

Since outliers predominantly occupy one half of each segment, we introduce \textbf{Half Rotation}---a low-cost strategy that couples dimensions $d^j_{i}$ and $d^j_{i+D_j/2}$ for $i \in \{1, \dots, D_j/2\}$. Here, $d^j_{i}$ represents the $i$-th dimension of segment $j \in \{f, h, w\}$.

We define a block-diagonal permutation matrix $\mathbf{\Pi} = \mathrm{diag}(\mathbf{\Pi}^f, \mathbf{\Pi}^h, \mathbf{\Pi}^w)$, where each $\mathbf{\Pi}^j$ reorders the dimensions of segment $j$ such that dimensions separated by $D_j/2$ are moved to adjacent positions. The rotation matrix $\mathbf{R}$ is similarly structured as $\mathbf{R} = \mathrm{diag}(\mathbf{R}^f, \mathbf{R}^h, \mathbf{R}^w)$, where each $\mathbf{R}^j$ consists of $D_j/2$ orthogonal blocks $\mathbf{R}^j_k \in \mathrm{O}(2)$.

The final rotated query matrix is obtained via segment-wise transformation and concatenation:
\begin{equation}
    \mathbf{Q}\mathbf{\Pi}\mathbf{R} = \left[ \mathbf{Q}_f \mathbf{\Pi}^f \mathbf{R}^f, \mathbf{Q}_h \mathbf{\Pi}^h \mathbf{R}^h, \mathbf{Q}_w \mathbf{\Pi}^w \mathbf{R}^w \right].
\end{equation}

For each segment, the operation $\mathbf{Q}_j(\mathbf{\Pi}^j \mathbf{R}^j)$ exploits the inherent sparsity of the combined permutation-rotation matrix. Consequently, this matrix multiplication can be transformed into a highly efficient element-wise operation, analogous to the optimized implementation of RoPE. Through Half Rotation, incoherence is effectively equalized across the entire segment while maintaining high computational efficiency.

Fig.~\ref{fig:rotation_strategies} summarizes the two proposed RoPE-aware rotation strategies. Interleaved Rotation groups adjacent channel dimensions and can be merged into RoPE, while Half Rotation pairs dimensions across the two half-segments to equalize RoPE-induced outliers with negligible overhead.

\subsubsection{Learnable Rotation (Preliminary)}
\label{sssec:learnable_rotation}

As a preliminary investigation, we also explore making the rotation \textbf{learnable}. Let $\mathbf{Y} = \mathbf{Q}^{(s)}(\mathbf{K}^{(s)})^\top$ denote the full-precision attention score matrix computed from the smoothed query and key matrices. We learn the rotation matrix $\mathbf{R}$ by minimizing the following calibration objective with gradient-based optimization:
\begin{align}
\mathbf{R}^{\star}
&=
\arg\min_{\{\mathbf{R}_i\}}
\left\|
\mathbf{Y}
-
\mathcal{Q}\!\left(\mathbf{Q}^{(s)}\mathbf{R}\right)
\mathcal{Q}\!\left((\mathbf{K}^{(s)}\mathbf{R})^\top\right)
\right\|_F^2,
\label{eq:learnable_rotation_obj}
\\
&\quad\text{s.t.}\quad
\mathbf{R}_i \in \mathrm{O}(2), \quad i=1,\dots,D/2 .
\end{align}
Here, $\mathbf{Q}^{(s)}$ and $\mathbf{K}^{(s)}$ are the smoothed matrices, and $\mathcal{Q}(\cdot)$ denotes the quantization function. As shown in \cref{sssec:orthogonal_property}, this orthogonality constraint is essential---without it, the rotation degrades attention output quality, confirming that LLM-style unconstrained rotations~\cite{liu2025spinquant,sun2025flatquant} do not transfer to 
mixed-precision INT4 attention in 3D-RoPE video DiTs.
This experiment serves primarily to validate the orthogonality requirement; a thorough calibration study is left to future work (see \cref{sec:experiments}).

\subsection{Range-optimized $\mathbf{P}$ Quantization}

In the FlashAttention framework \cite{dao2022flashattention,dao2024flashattention,shah2024flashattention}, the attention scores are computed block-wise. Let $\mathbf{S}_{i,j} = \mathbf{Q}_i \mathbf{K}_j^\top / \sqrt{D}$ denote the pre-softmax attention scores for a given block. To maintain numerical stability during the online softmax process, a running row maximum $m_{i,j} = \max(m_{i,j-1}, \text{rowmax}(\mathbf{S}_{i,j}))$ is tracked, and the local attention probabilities are defined as $\mathbf{P}_{i,j} = \exp(\mathbf{S}_{i,j} - m_{i,j})$.

\subsubsection{Affine Range Expansion}

Standard quantization practices typically normalize the attention probabilities by their maximum value to fit the target bit-width. For $\text{INT4}$ symmetric quantization, the mapping is:
\begin{equation}
    \tilde{\mathbf{P}}_{i,j}^{\text{sym}} = \left\lfloor \frac{\mathbf{P}_{i,j}}{\max(\mathbf{P}_{i,j})} \times 7 \right\rceil \in (0, 7].
\end{equation}
This approach, however, leaves the negative representative space of $\text{INT4}$ ($[-8, 0)$) entirely unused, effectively wasting half of the available quantization levels and leading to significant precision loss.

To maximize the utilization of the 4-bit dynamic range, we propose a mapping with a fixed scale of $15$ and a zero-point of $-8$. This affine transformation projects the normalized probabilities onto the full $\text{INT4}$ interval $[-8, 7]$:
\begin{equation}
    \tilde{\mathbf{P}}_{i,j} = \left( \frac{\mathbf{P}_{i,j}}{\max(\mathbf{P}_{i,j})} \times 15 \right) - 8 \in (-8, 7].
\end{equation}
By expanding the representation from 8 levels ($0$ to $7$) to 16 levels ($-8$ to $7$), we effectively double the quantization resolution, significantly reducing the rounding error.

\subsection{Hadamard Transformation for the Value Matrix}
\label{ssec:hadamard_v}

To mitigate the impact of outliers within the Value matrix $\mathbf{V}$, we employ an orthogonal rotation transformation using a $D \times D$ Hadamard matrix $\mathbf{H}$, where $D = 2^n$ ($n \in \mathbb{Z}^+$). The $2 \times 2$ Hadamard matrix is defined in \cref{eq:H_2}. Hadamard matrices are constructed recursively via the Kronecker product:
\begin{equation}
\mathbf{H}_{2^{n}} = \mathbf{H}_{2} \otimes \mathbf{H}_{2^{n-1}}.
\end{equation}

The Hadamard rotation can be mathematically absorbed into the value projection $\mathbf{W}_v$ and the output projection $\mathbf{W}_o$. Given the value projection $\mathbf{V} = \mathbf{X}\mathbf{W}_v$, the standard attention output is:

\begin{align}
\mathbf{O} &= \text{Softmax}\left(\frac{\mathbf{Q}\mathbf{K}^\top}{\sqrt{d_k}}\right) \mathbf{V} \mathbf{W}_o \nonumber \\
 &= \text{Softmax}\left(\frac{\mathbf{Q}\mathbf{K}^\top}{\sqrt{d_k}}\right) \mathbf{X} (\mathbf{W}_v \mathbf{H}) (\mathbf{H}^\top \mathbf{W}_o).
\end{align}

\newpage

We can merge the Hadamard transform into the weights as $\widetilde{\mathbf{W}}_v=\mathbf{W}_v\mathbf{H}$ and $\widetilde{\mathbf{W}}_o=\mathbf{H}^{\top}\mathbf{W}_o$, respectively. Since these modified weights are computed offline during the preprocessing stage, the Hadamard transformation for $\mathbf{V}$ is achieved with \textbf{zero additional computational overhead} during inference.

\section{Experiments}
\label{sec:experiments}

\subsection{Experimental Setup}
We evaluate our proposed method on the \textbf{Wan2.2}~\cite{wan2025wan} T2V and I2V models at 480P resolution with 81 frames (40 sampling steps), and the \textbf{HunyuanVideo}~\cite{kong2024hunyuanvideo} T2V model at 480P resolution with 129 frames (50 sampling steps). Our evaluation dataset comprises diverse prompts curated from the official Wan2.2, Wan-Lightning~\cite{fan2025wanlighting}, and Open-Sora~\cite{zheng2024opensora} example sets, following the evaluation protocol of SageAttention~\cite{zhang2025sageattention1}. For the attention mechanism, we employ \textbf{per-token quantization} for the query ($\mathbf{Q}$) and key ($\mathbf{K}$) matrices, and \textbf{per-group quantization} for the value ($\mathbf{V}$) matrix.

For \textbf{Wan2.2-I2V}, we preserve FP16 precision for the first 6 and last 4 sampling steps, as well as the first and last 2 DiT blocks; all remaining FlashAttention operations are executed with our INT4 attention kernel, resulting in 68\% of FlashAttention operations running in INT4. For \textbf{Wan2.2-T2V}, FP16 is maintained for the first and last 2 sampling steps and DiT blocks, enabling 81\% INT4 attention execution. Similarly, for \textbf{HunyuanVideo-T2V}, we keep FP16 for the first 4 and last 2 sampling steps and the first and last 2 DiT blocks, with 76\% of attention operations running in INT4.

\paragraph{\textbf{Hardware Configuration.}} We benchmark INT4 attention kernel speed on \textbf{NVIDIA A10} GPUs using custom CUDA kernels, reflecting practical deployment hardware. For accuracy evaluation of the full video generation pipeline, we use \textbf{NVIDIA H20} GPUs with \textbf{TileLang}~\cite{wang2025tilelang} operators, which provide precise numerical validation.

\paragraph{\textbf{Model Scope.}} We evaluate on Wan2.2 and HunyuanVideo, two leading open-source video generation models, both employing DiT-based architectures with 3D RoPE. This choice is deliberate: DiT with 3D RoPE has become the \emph{de facto} standard for state-of-the-art open-source video generation (also adopted by CogVideoX~\cite{yang2024cogvideox}, Open-Sora 2.0~\cite{peng2025open}, and others), making it the most practically relevant setting. 

\subsection{Evaluation Metrics}
We primarily employ \textbf{Relative Difference Metrics} to quantify the discrepancy between our quantized models and the FP16 baseline. Specifically, \textbf{PSNR}~\cite{korhonen2012peak} and \textbf{Cosine Similarity} measure low-level pixel-space differences, while \textbf{SSIM}~\cite{wang2004ssim} evaluates structural similarity. These relative metrics are more sensitive to the fine-grained artifacts introduced by quantization~\cite{zhao2025paroattention}, providing a clearer signal for comparing quantization strategies.

We also evaluate absolute video quality using \textbf{VBench}~\cite{huang2024vbench}, which assesses semantic coherence, temporal consistency, and motion quality. Since VBench scores are less discriminative across quantization variants (methods with visible pixel-level differences often receive similar VBench scores), we use relative metrics as the primary comparison in the main paper and report VBench results in the supplementary material.

\subsection{Baselines and Ablation Study}
The primary baseline is the standard attention mechanism in \textbf{FP16}. We also compare against \textbf{SageAttention}~\cite{zhang2025sageattention1} with $\mathbf{Q}\mathbf{K}\mathbf{V}$ smoothing but without our optimized $\mathbf{P}$ and rotation strategies. Furthermore, we evaluate a full rotation baseline using a $D \times D$ Hadamard matrix. Extensive \textbf{ablation studies} are performed to analyze the impact of different rotation and optimized $\mathbf{P}$ strategies. For the Half Rotation and Interleaved Rotation results in Table~\ref{tab:quality_comparison}, the rotation matrix $\mathbf{R}$ defaults to the block-diagonal initialization using $\mathbf{H}_2$.

When applying rotations to $\mathbf{Q}$ and $\mathbf{K}$, we include a Hadamard transformation on $\mathbf{V}$ by default. Since this transformation is computationally mergeable---and follows standard practice in frameworks like FlatQuant and SpinQuant~\cite{liu2025spinquant,sun2025flatquant}---we adopt it as a default component and omit separate ablation studies for this operation.

We note that PAROAttention~\cite{zhao2025paroattention} addresses the complementary problem of token reordering for $\mathbf{P}$ quantization, which operates at the token level rather than the channel level. Our rotation strategy and PAROAttention's reordering are orthogonal techniques that can be combined. We provide a software-level comparison and combination experiment in the supplementary material, demonstrating additive accuracy gains when both approaches are applied.

\paragraph{\textbf{Learned Rotation Setup.}}
As a preliminary investigation into learned rotations (see \cref{sssec:learnable_rotation}), we train under two conditions: \textbf{(1) Unconstrained} and \textbf{(2) Orthogonal}. We use Half Rotation with $\mathbf{H}_2$ initialization on Wan2.2-I2V and 1--2 calibration samples.

\subsection{Experimental Results}

Table~\ref{tab:quality_comparison} summarizes the impact of different rotation strategies on the Wan2.2 (I2V/T2V) and HunyuanVideo models.

\paragraph{\textbf{Quantitative Analysis.}}
Difference metrics indicate that \textbf{Half Rotation} achieves the best performance for Wan2.2-I2V in terms of \textbf{Cosine Similarity} and \textbf{MSE}, while remaining highly competitive in \textbf{SSIM} and \textbf{PSNR}. On HunyuanVideo, Half Rotation yields the best Cosine Similarity, MSE, and SSIM. Notably, combining Half Rotation with \textbf{Optimized $\mathbf{P}$ Quantization} consistently outperforms SageAttention across all tasks.

\begin{table*}[!p]
  \centering
  \scriptsize
  \caption{Difference Metrics for Wan2.2 I2V, T2V and HunyuanVideo T2V}
  \label{tab:quality_comparison}
  \newcolumntype{C}{>{\centering\arraybackslash}X}
  
  \begin{tabularx}{\linewidth}{l l c l C C C C c}
    \toprule
    \multirow{2}{*}{\textbf{Model}} & \multicolumn{3}{c}{\textbf{Method}} & \multicolumn{4}{c}{\textbf{Difference Metrics}} & \multirow{2}{*}{\textbf{Speedup}} \\
    \cmidrule(lr){2-4} \cmidrule(lr){5-8}
    & Datatype & Optim-P & Rotation & Cosine $\uparrow$ & MSE $\downarrow$ & SSIM $\uparrow$ & PSNR $\uparrow$ & \\
    \midrule
    \multirow{7}{*}{\begin{tabular}{@{}l@{}}Wan2.2 \\ -I2V\end{tabular}} 
    & \cellcolor{gray!20}FP16 & -- & -- & 1.0000 & 0.0000 & 1.0000 & $\infty$ & -- \\
    & INT4 & $\times$ & W/O & 0.9809 & 519.30 & 0.7752 & 22.828 & \multirow{6}{*}{1.51$\times$} \\
    & INT4 & \checkmark & W/O & 0.9789 & 575.17 & 0.7524 & 22.069 & \\
    & INT4 & \checkmark & Half & \textbf{0.9824} & \textbf{493.35} & 0.7767 & 22.903 & \\
    & INT4 & \checkmark & Interleave & 0.9792 & 583.88 & 0.7502 & 22.057 & \\
    & INT4 & \checkmark & Full & 0.9807 & 515.34 & \textbf{0.7808} & \textbf{23.176} & \\
    & INT4 & \checkmark & Trained & 0.9809 & 519.94 & 0.7746 & 22.713 & \\
    \midrule
    \multirow{6}{*}{\begin{tabular}{@{}l@{}}Wan2.2 \\ -T2V\end{tabular}} 
    & \cellcolor{gray!20}FP16 & -- & -- & 1.0000 & 0.0000 & 1.0000 & $\infty$ & -- \\
    & INT4 & $\times$ & W/O & 0.9458 & 1574.5 & 0.6348 & 17.829 & \multirow{5}{*}{1.68$\times$} \\
    & INT4 & \checkmark & W/O & 0.95178 & 1410.1 & 0.6582 & 18.427 & \\
    & INT4 & \checkmark & Half & 0.9492 & 1466.0 & 0.6504 & 18.462 & \\
    & INT4 & \checkmark & Interleave & \textbf{0.95622} & 1392.1 & 0.6628 & 18.544 & \\
    & INT4 & \checkmark & Full & 0.95270 & \textbf{1370.2} & \textbf{0.6737} & \textbf{19.211} & \\
    \midrule
    \multirow{6}{*}{\begin{tabular}{@{}l@{}}Hyvideo \\ -T2V\end{tabular}} 
    & \cellcolor{gray!20}FP16 & -- & -- & 1.0000 & 0.0000 & 1.0000 & $\infty$ & -- \\
    & INT4 & $\times$ & W/O & 0.9663 & 1095.3 & 0.6621 & 18.332 & \multirow{5}{*}{1.61$\times$} \\
    & INT4 & \checkmark & W/O & 0.9712 & 1089.6 & 0.6620 & 18.945 & \\
    & INT4 & \checkmark & Half & \textbf{0.9751} & \textbf{1041.9} & \textbf{0.6677} & 18.971 & \\
    & INT4 & \checkmark & Interleave & 0.9750 & 1045.8 & 0.6588 & \textbf{19.037} & \\
    & INT4 & \checkmark & Full & 0.9744 & 1076.4 & 0.6646 & 18.854 & \\
    \bottomrule
  \end{tabularx}
\end{table*}

\begin{figure*}[!p]

  \centering
  \captionsetup{skip=2pt}
  \newcommand{\myimgwidth}{0.14\linewidth}
  \setlength{\tabcolsep}{0pt}
  {\scriptsize
  \begin{tabular}{cccccc}
     (a) FP16 & (b) Full & (c) Half & (d) Interleave & (e) W/O Rot & (f) SageAttn \\
    \includegraphics[width=\myimgwidth]{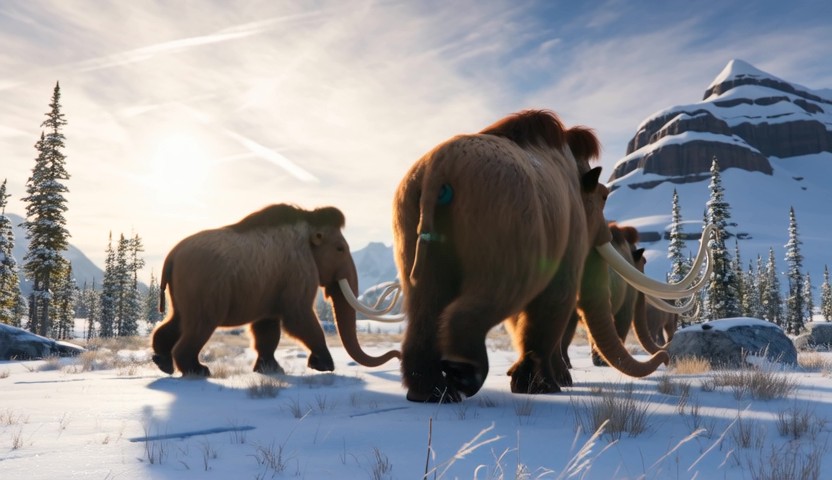} &
    \includegraphics[width=\myimgwidth]{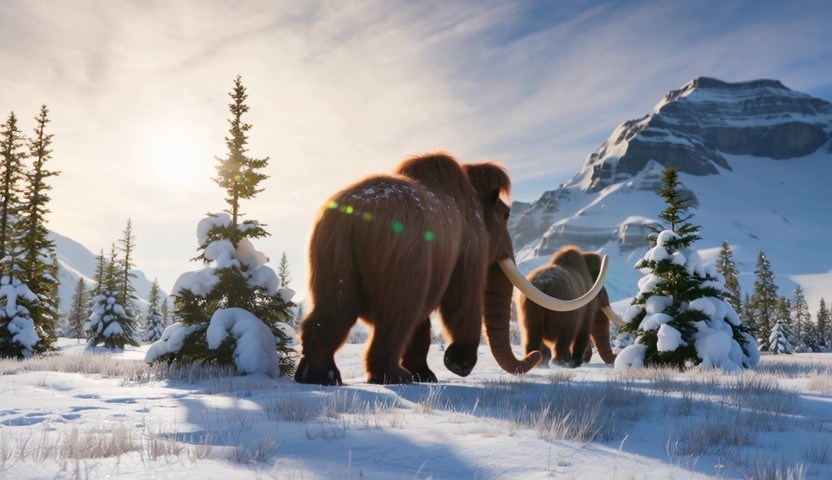} &
    \includegraphics[width=\myimgwidth]{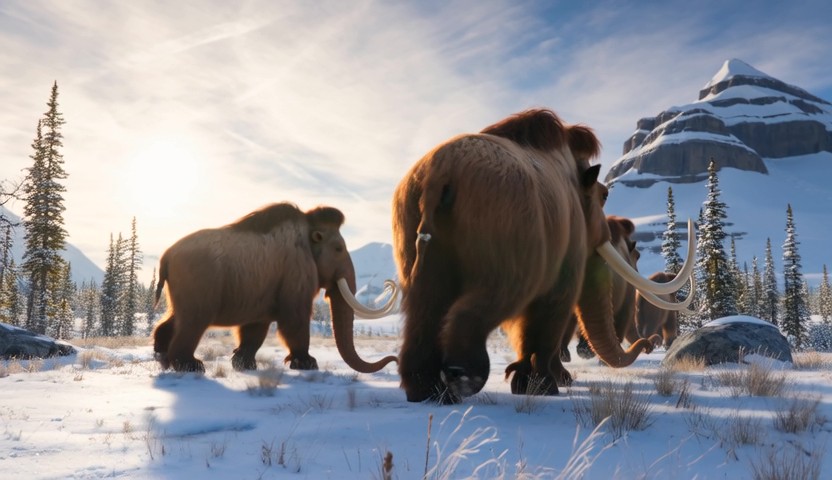} &
    \includegraphics[width=\myimgwidth]{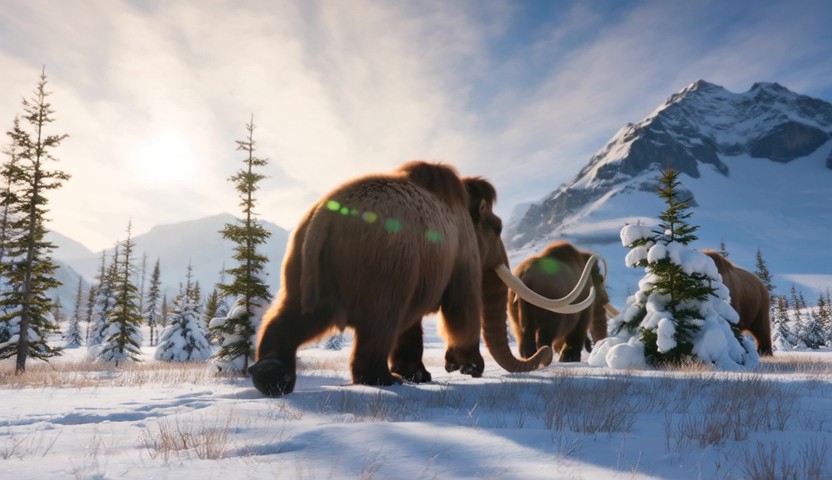} &
    \includegraphics[width=\myimgwidth]{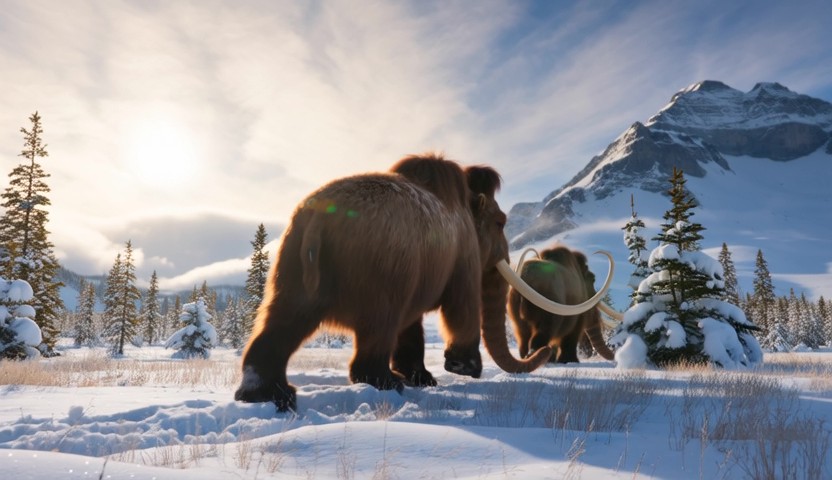} &
    \includegraphics[width=\myimgwidth]{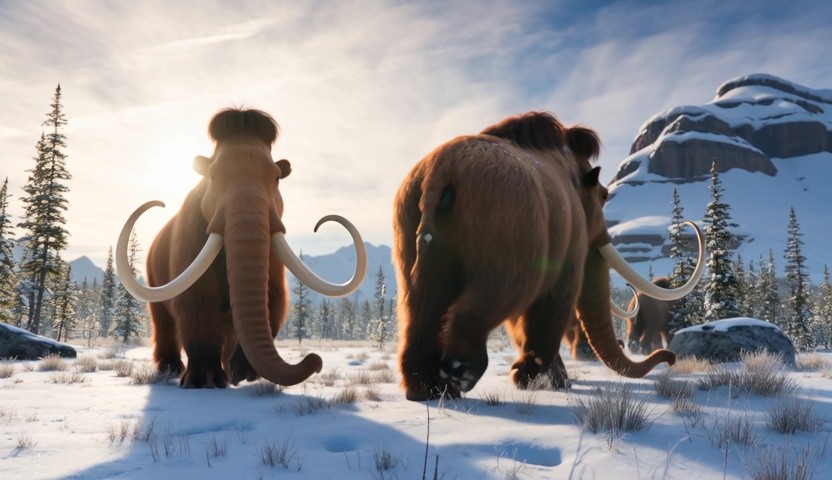} \\

    \includegraphics[width=\myimgwidth]{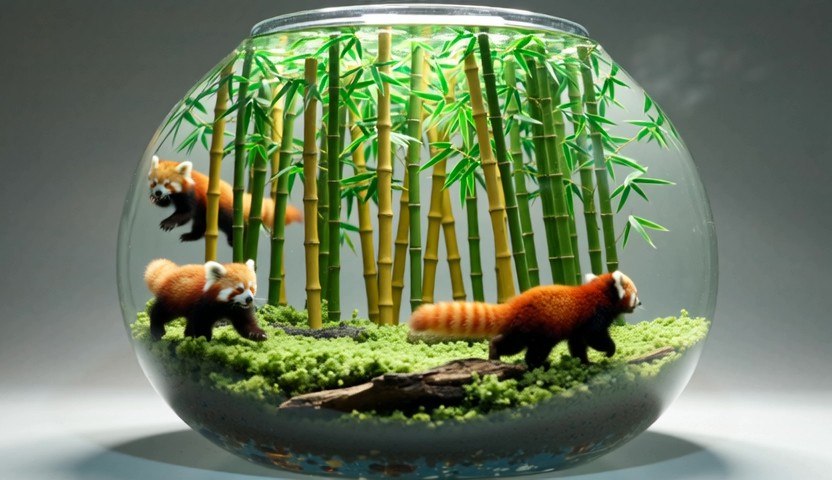} &
    \includegraphics[width=\myimgwidth]{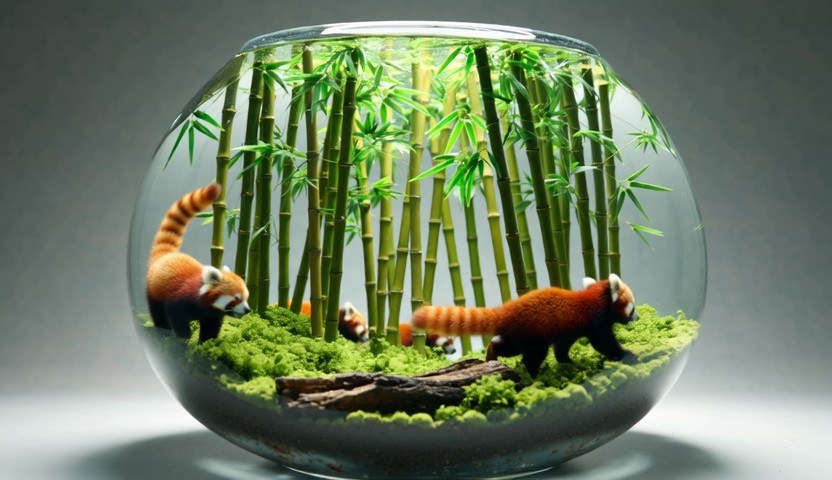} &
    \includegraphics[width=\myimgwidth]{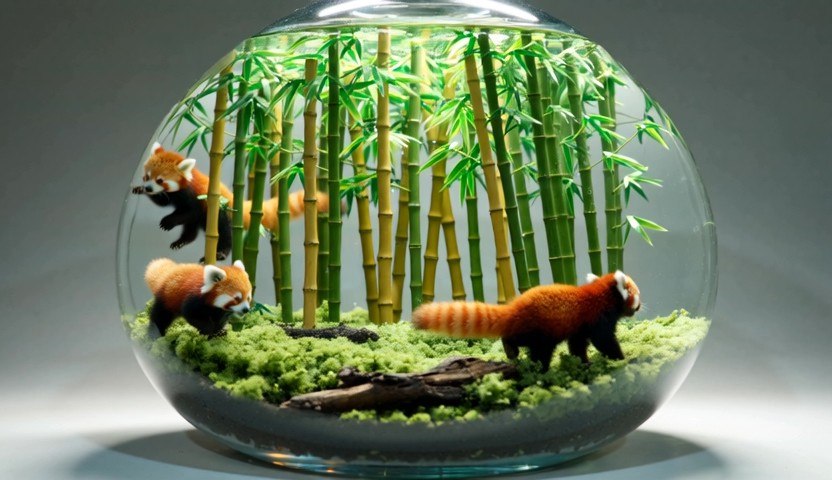} &
    \includegraphics[width=\myimgwidth]{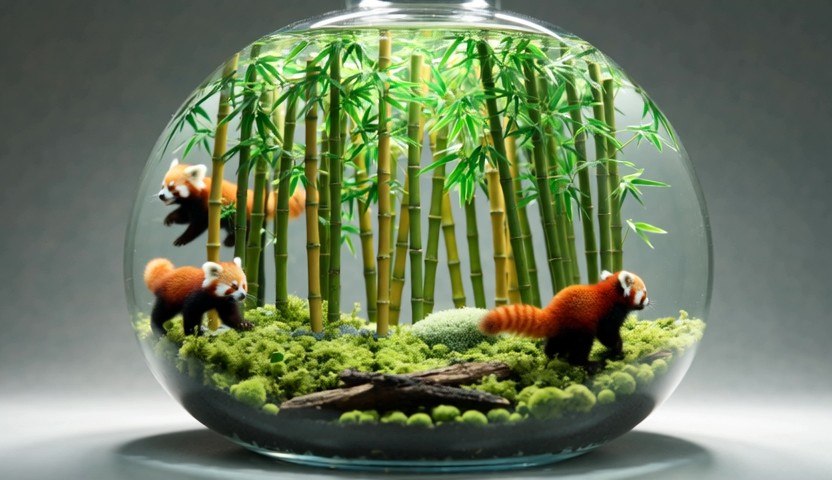} &
    \includegraphics[width=\myimgwidth]{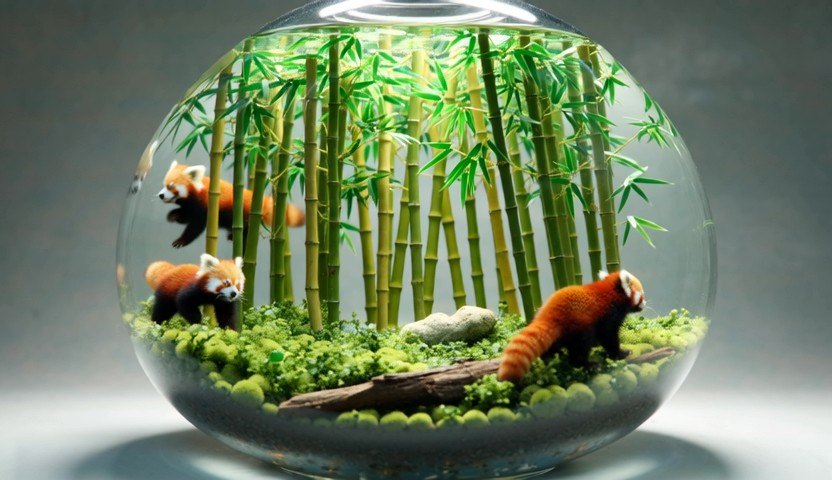} &
    \includegraphics[width=\myimgwidth]{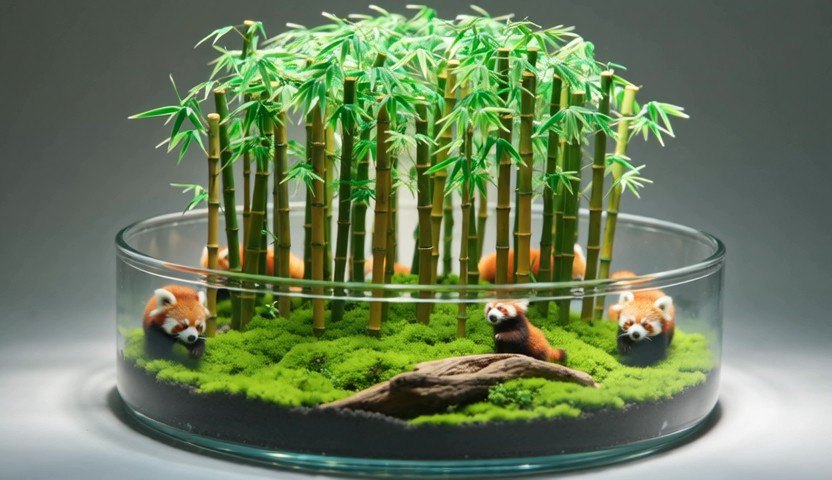} \\

    \includegraphics[width=\myimgwidth]{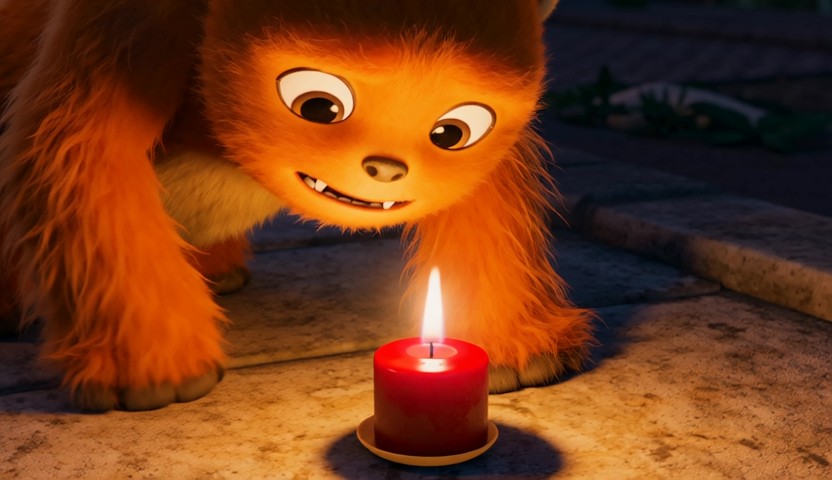} &
    \includegraphics[width=\myimgwidth]{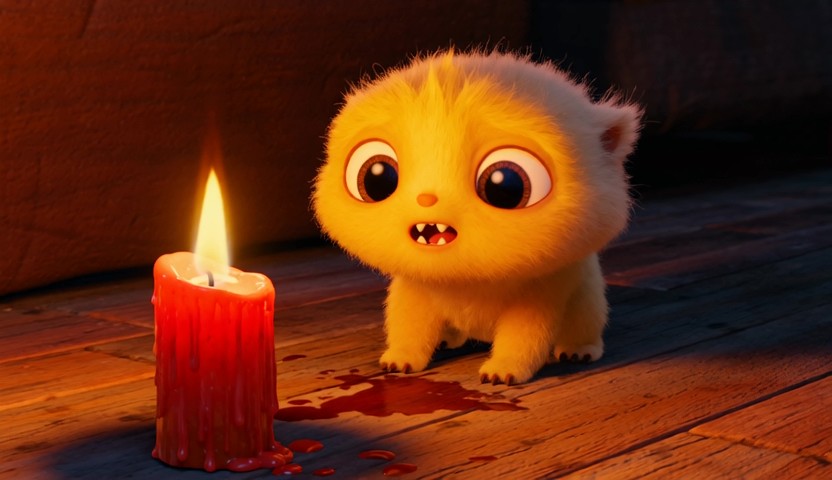} &
    \includegraphics[width=\myimgwidth]{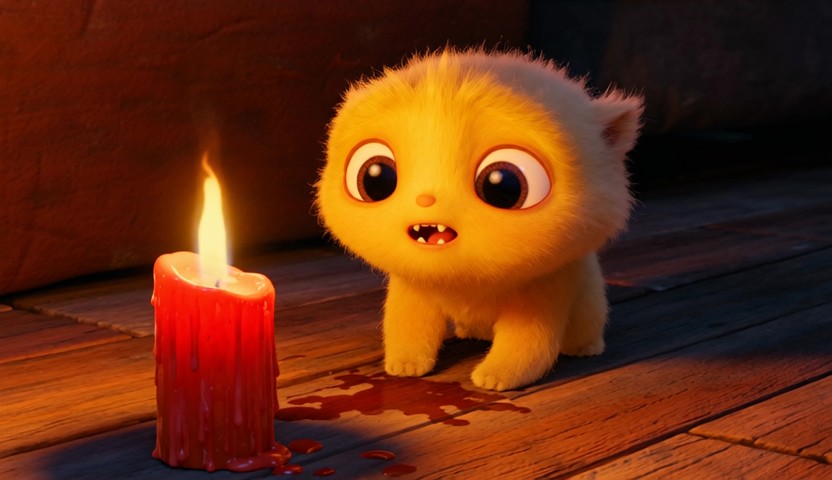} &
    \includegraphics[width=\myimgwidth]{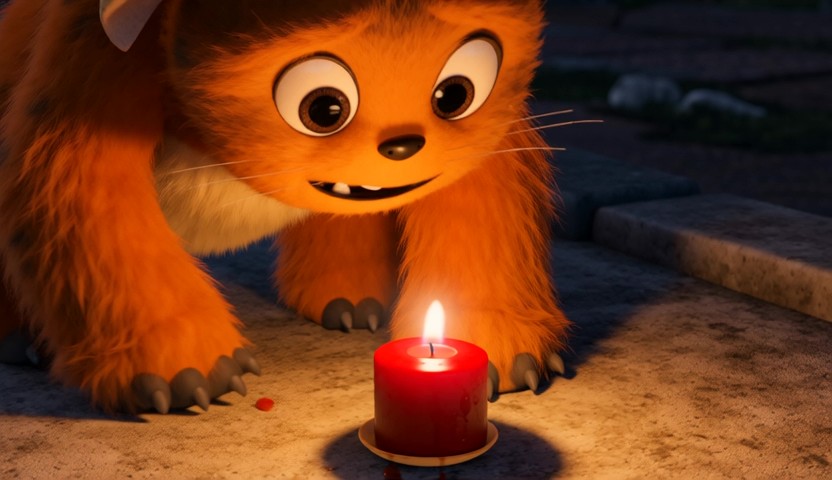} &
    \includegraphics[width=\myimgwidth]{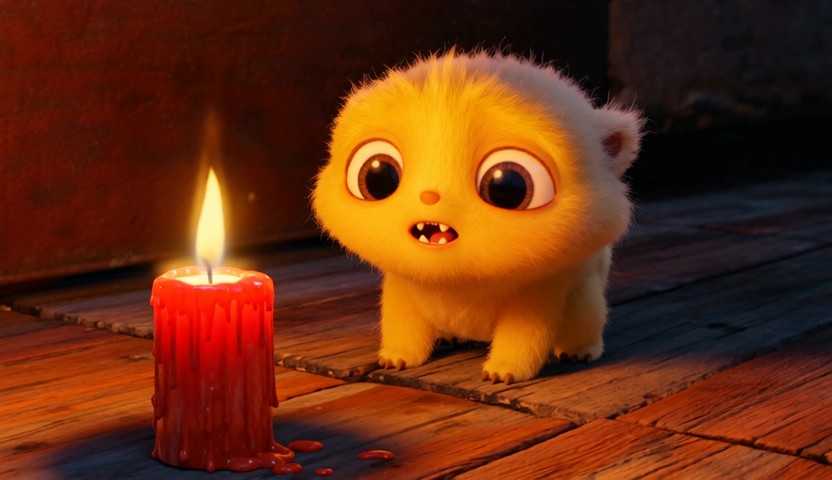} &
    \includegraphics[width=\myimgwidth]{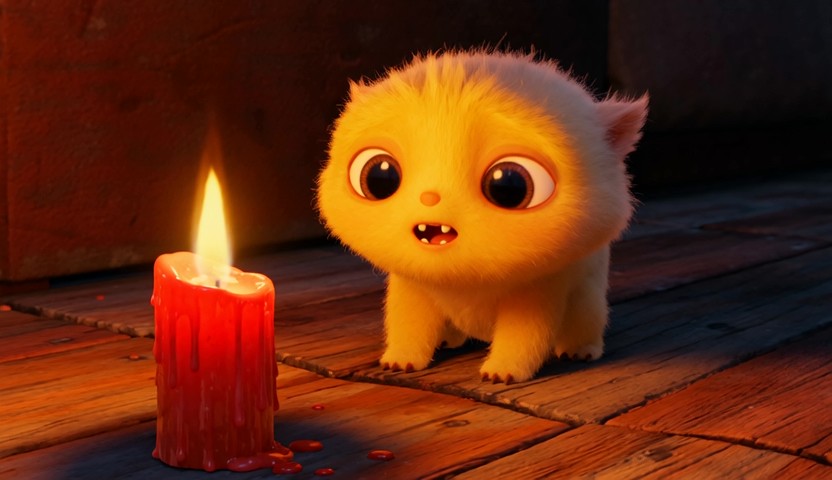} \\
  \end{tabular}
  }
  \caption{Comparison of different rotation strategies in Wan2.2 T2V models.}
  \label{fig:wan2_2_t2v}
\end{figure*}

\begin{figure*}[!p]
  \centering
  \captionsetup{skip=1pt}
  \newcommand{\myimgwidth}{0.27\linewidth}
  \setlength{\tabcolsep}{1pt}
  {\scriptsize
  \begin{tabular}{ccc}
    \includegraphics[width=\myimgwidth]{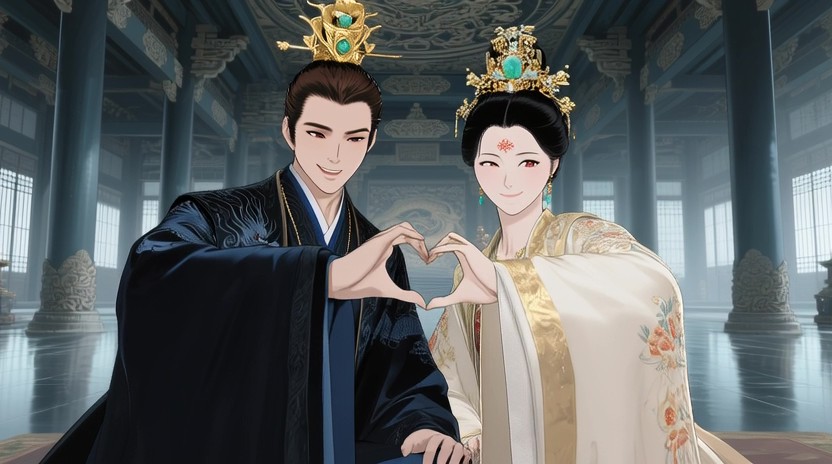} &
    \includegraphics[width=\myimgwidth]{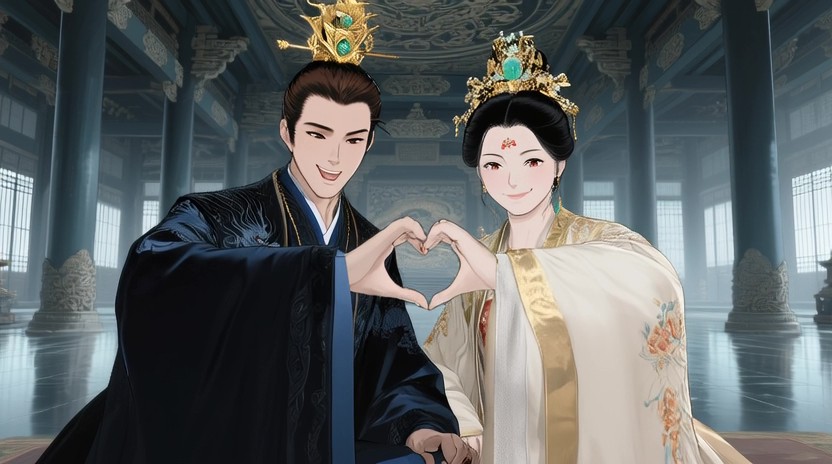} &
    \includegraphics[width=\myimgwidth]{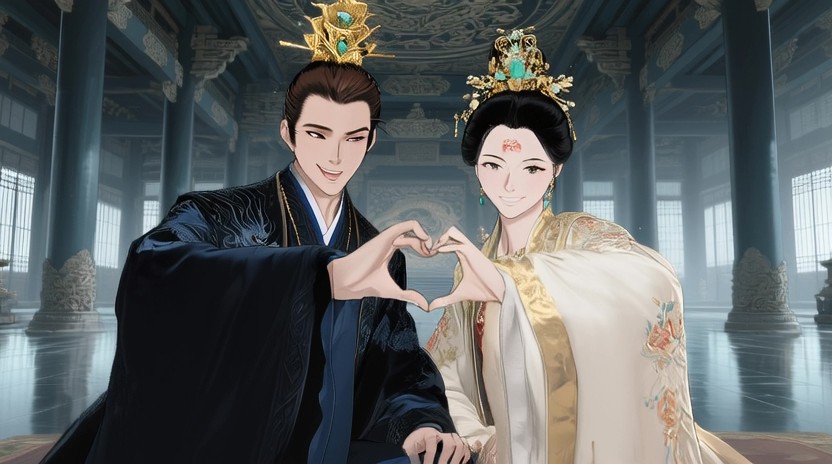} \\[1pt]

    \includegraphics[width=\myimgwidth]{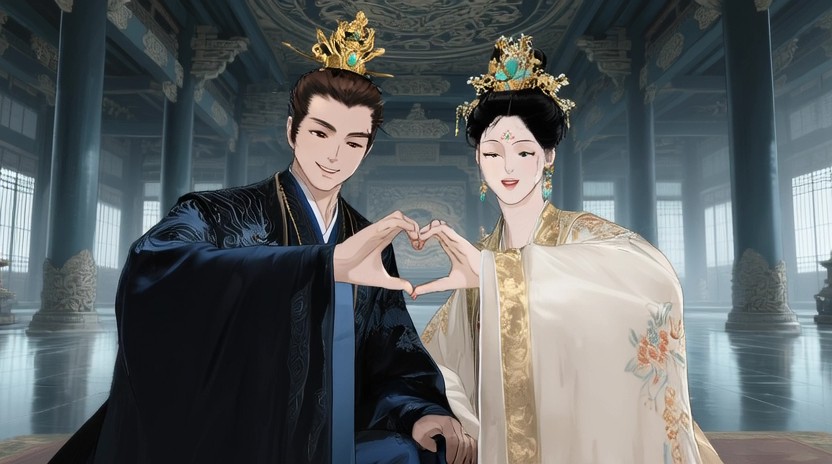} &
    \includegraphics[width=\myimgwidth]{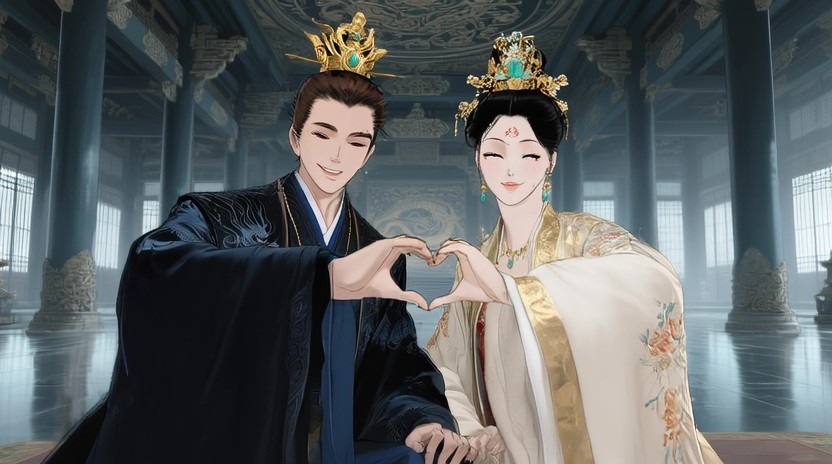} &
    \includegraphics[width=\myimgwidth]{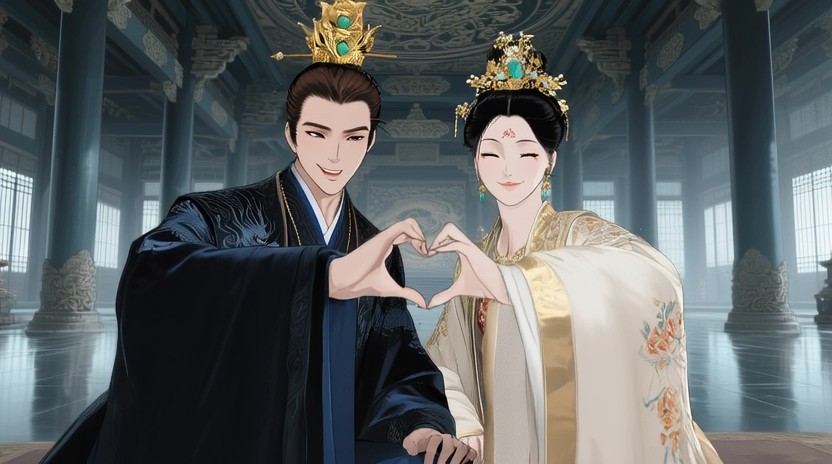} \\
  \end{tabular}
  }
  \caption{Comparison of different rotation strategies in Wan2.2 I2V models (top row: FP16/Full/Half; bottom row: Interleave/W/O Rot/SageAttn).}
  \label{fig:wan2_2_i2v}
\end{figure*}

\begin{figure*}[!t]
  \centering
  \captionsetup{skip=1pt}
  \newcommand{\myimgwidth}{0.27\linewidth}
  \setlength{\tabcolsep}{1pt}
  {\scriptsize
  \begin{tabular}{ccc}
    \includegraphics[width=\myimgwidth]{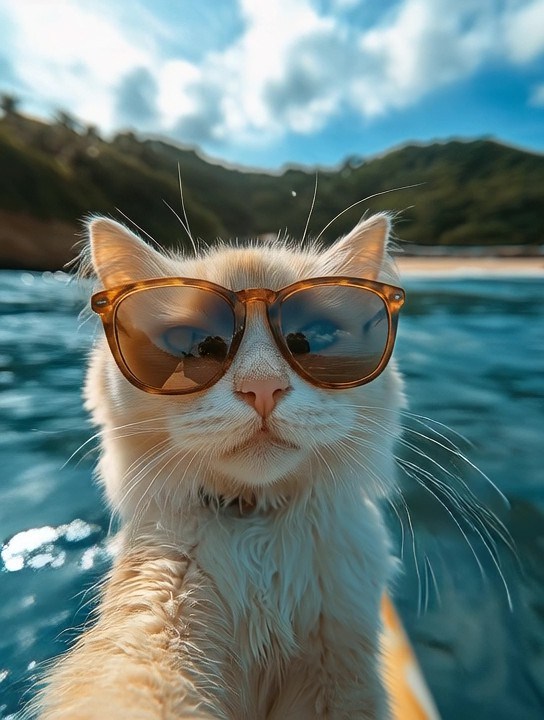} &
    \includegraphics[width=\myimgwidth]{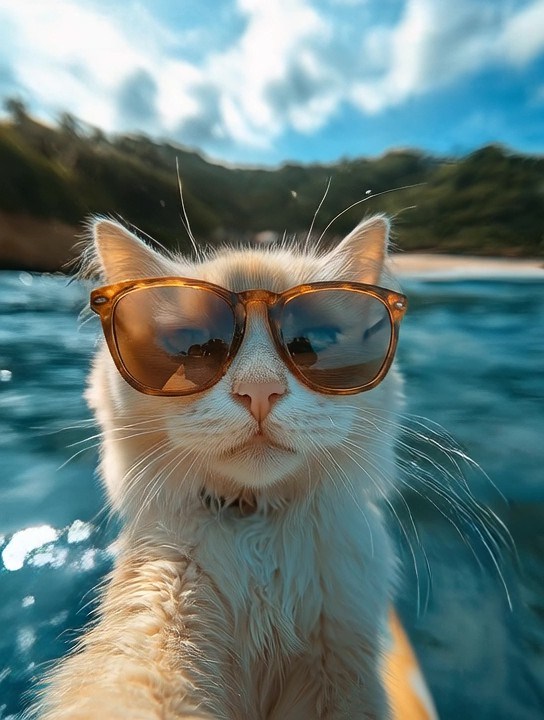} &
    \includegraphics[width=\myimgwidth]{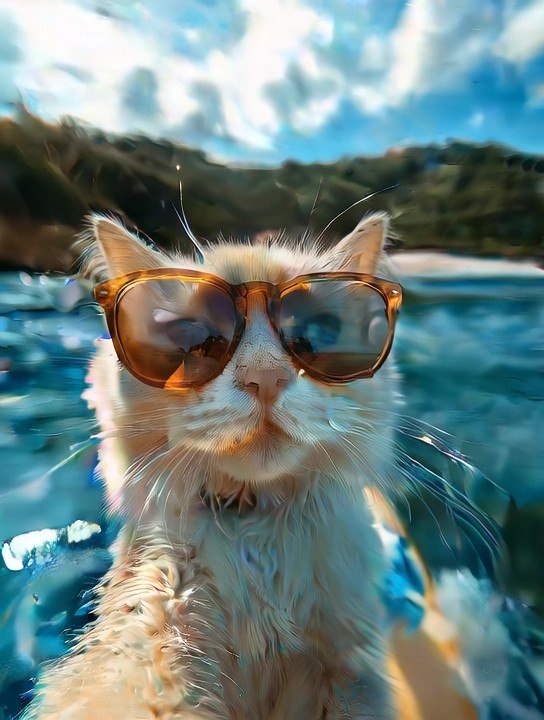} \\
  \end{tabular}
  }
  \caption{Comparison of effect of trained Rotation (left to right: FP16, trained with orthogonality, trained without orthogonality).}
  \label{fig:train}
\end{figure*}

The \textbf{Interleaved Rotation} method significantly improves Cosine Similarity for Wan2.2-I2V, though gains in other metrics are marginal. However, on Wan2.2-T2V, it achieves the highest Cosine Similarity and delivers substantial improvements in MSE, SSIM, and PSNR---second only to Full Rotation. For HunyuanVideo, this method achieves the best PSNR. While its performance on Wan2.2-I2V and HunyuanVideo is comparable to SageAttention, it demonstrates clear superiority on the Wan2.2-T2V model.

\textbf{Range-Optimized $\mathbf{P}$ Quantization} yields clear improvements on Wan2.2-T2V and HunyuanVideo across all metrics, yet provides marginal gains on Wan2.2-I2V. We attribute this to the I2V model's image conditioning, which constrains the attention distribution $\mathbf{P}$ to be sharply peaked (low entropy) with values concentrated near 0 and 1. In this regime, standard symmetric quantization effectively preserves the dominant structure, while range-optimized quantization can introduce subtle precision degradation. In contrast, T2V models produce more diffuse $\mathbf{P}$ distributions that benefit substantially from the full INT4 dynamic range. These findings suggest that range-optimized quantization is most beneficial for moderate-to-high entropy attention, and may be disabled for strongly conditioned, low-entropy patterns.

\paragraph{\textbf{Analysis of Rotation Strategy Differences.}}
Performance differences match RoPE-induced outlier structure: \textbf{Half Rotation} is strongest when outliers concentrate in one half-segment (I2V and HunyuanVideo), while \textbf{Interleaved Rotation} is better when local adjacent-channel correlation dominates (T2V). \textbf{Full Rotation} can over-smooth useful local structure despite stronger global redistribution.

\textbf{Practical recommendation:} For deployment on new models, we recommend \textbf{Half Rotation} as the default strategy due to its consistent performance across all tested configurations. Interleaved Rotation offers a compelling zero-cost alternative when it can be pre-fused into RoPE weights, particularly for T2V applications.

Visual inspection in Fig.~\ref{fig:wan2_2_t2v} reveals that for T2V tasks, Interleaved and Half Rotation occasionally outperform Full Rotation in specific cases. Moreover, as shown in Figs.~\ref{fig:wan2_2_t2v} and~\ref{fig:wan2_2_i2v}, both strategies exhibit clear visual superiority over the SageAttention baseline, particularly in maintaining structural integrity and temporal consistency.

\paragraph{\textbf{Learnable Rotation Result.}}
Learned rotation is effective \emph{only} under strict orthogonality constraints, as shown in Fig.~\ref{fig:train}; unconstrained training degrades quality (\cref{sssec:orthogonal_property}), confirming that LLM-style unconstrained rotations~\cite{liu2025spinquant,sun2025flatquant} do not transfer directly to 3D-RoPE video DiTs. With 1--2 calibration samples, orthogonal learned rotation does not outperform fixed Half Rotation.

\paragraph{\textbf{Speed and Cost Analysis.}} Our INT4 kernel yields 2.2$\times$ speedup versus custom FP16 attention, and end-to-end speedups of 1.51$\times$ (Wan2.2-I2V), 1.68$\times$ (Wan2.2-T2V), and 1.61$\times$ (HunyuanVideo). Full rotation scales linearly with sequence length, whereas attention scales quadratically. Consequently, for sequences exceeding 30k tokens, full rotation overhead remains under 1\% of attention computation. However, when paired with sparse attention on shorter sequences ($<$10k tokens), its relative overhead increases to over 10\% of the attention computation. Notably, Half Rotation reduces this overhead to approximately 3\% of the full rotation cost.

\section{Conclusion}

In this work, we propose \textbf{RotateAttention}, an efficient mixed-precision INT4 attention framework designed for DiT-based video generation models with 3D RoPE.
Our approach introduces two core innovations: 
(1) \textbf{RoPE-aware Rotation}, which utilizes mergeable or negligible-cost rotation matrices to effectively mitigate $\mathbf{Q}$ and $\mathbf{K}$ outliers while maintaining seamless integration with RoPE; 
(2) \textbf{Range-optimized $\mathbf{P}$ Quantization}, which leverages fixed scales and zero-points to fully exploit the \textbf{INT4 numerical range} with minimal computational overhead. Additionally, we apply a standard Hadamard transform to $\mathbf{V}$, fused offline into the linear projections at zero inference cost.

Experimental results demonstrate that \textbf{RotateAttention} achieves video generation quality nearly identical to full-precision baselines and significantly outperforms SageAttention in visual fidelity. Our INT4 attention kernel achieves a $2.2\times$ speedup at the kernel level, translating to end-to-end speedups of $1.51\times$--$1.68\times$ across the evaluated 3D-RoPE video DiTs. \textbf{RotateAttention} establishes a practical paradigm for high-performance, mixed-precision INT4 attention in 3D-RoPE video DiTs.

\paragraph{\textbf{Limitations.}} Our rotation strategies are specifically designed for DiT-based video generation models employing 3D RoPE with segmented dimensional partitioning. Applying them to other positional encoding schemes requires adapting the rotation structure to the corresponding outlier patterns. Our evaluation covers two model families, Wan2.2 and HunyuanVideo; evaluating a broader range of architectures would further strengthen generalizability.

%
%
\newpage
\bibliographystyle{splncs04}
\bibliography{main}

\end{document}